\pdfoutput=1
\documentclass[11pt]{article}

\usepackage[]{ACL2023}

\usepackage{times}
\usepackage{latexsym}
\usepackage{amsmath, mathtools}
\usepackage{booktabs}
\usepackage{makecell}

\usepackage[T1]{fontenc}
\usepackage[utf8]{inputenc}
\usepackage{float}

\usepackage{microtype}
\usepackage{amsfonts,amssymb}

\usepackage{inconsolata}
\usepackage{verbatim}
\usepackage{graphicx}
\usepackage{hyperref}
\definecolor{c1}{HTML}{ae7000}
\definecolor{c2}{HTML}{1685a9}
\definecolor{c3}{HTML}{057748}
\definecolor{c4}{HTML}{f20c00}

\title{A Cognitive Stimulation Dialogue System with Multi-source Knowledge Fusion for Elders with Cognitive Impairment}

\author{
Jiyue Jiang, 
Sheng Wang, 
Qintong Li,
\textbf{Lingpeng Kong},
\textbf{Chuan Wu}
\\
The University of Hong Kong \\
\texttt{\{jiangjy,u3009618,qtli\}@connect.hku.hk} \\ 
\texttt{\{lpk,cwu\}@cs.hku.hk }
}

\begin{document}
\maketitle
\begin{abstract}
When communicating with elders with cognitive impairment, cognitive stimulation (CS) help to maintain the cognitive health of elders. Data sparsity is the main challenge in building CS-based dialogue systems, particularly in the Chinese language.  To fill this gap, we construct a Chinese CS conversation (CSConv) dataset, which contains about 2.6K groups of dialogues with CS principles and emotional support strategy labels. Making chit chat while providing emotional support is overlooked by the majority of existing cognitive dialogue systems. In this paper, we propose a multi-source knowledge fusion method for CS dialogue (CSD), to generate open-ended responses guided by the CS principle and emotional support strategy. We first use a progressive mask method based on external knowledge to learn encoders as effective classifiers, which is the prerequisite to predict the CS principle and emotional support strategy of the target response. Then a decoder interacts with the perceived CS principle and emotional support strategy to generate responses. Extensive experiments conducted on the CSConv dataset demonstrate the effectiveness of the proposed method, while there is still a large space for improvement compared to human performance\footnote{Our data and code could be found in \url{https://github.com/jiangjyjy/CSD}}.
\end{abstract}

\section{Introduction}

\begin{figure}[h]
    \centering
    \includegraphics[width=7.5cm]{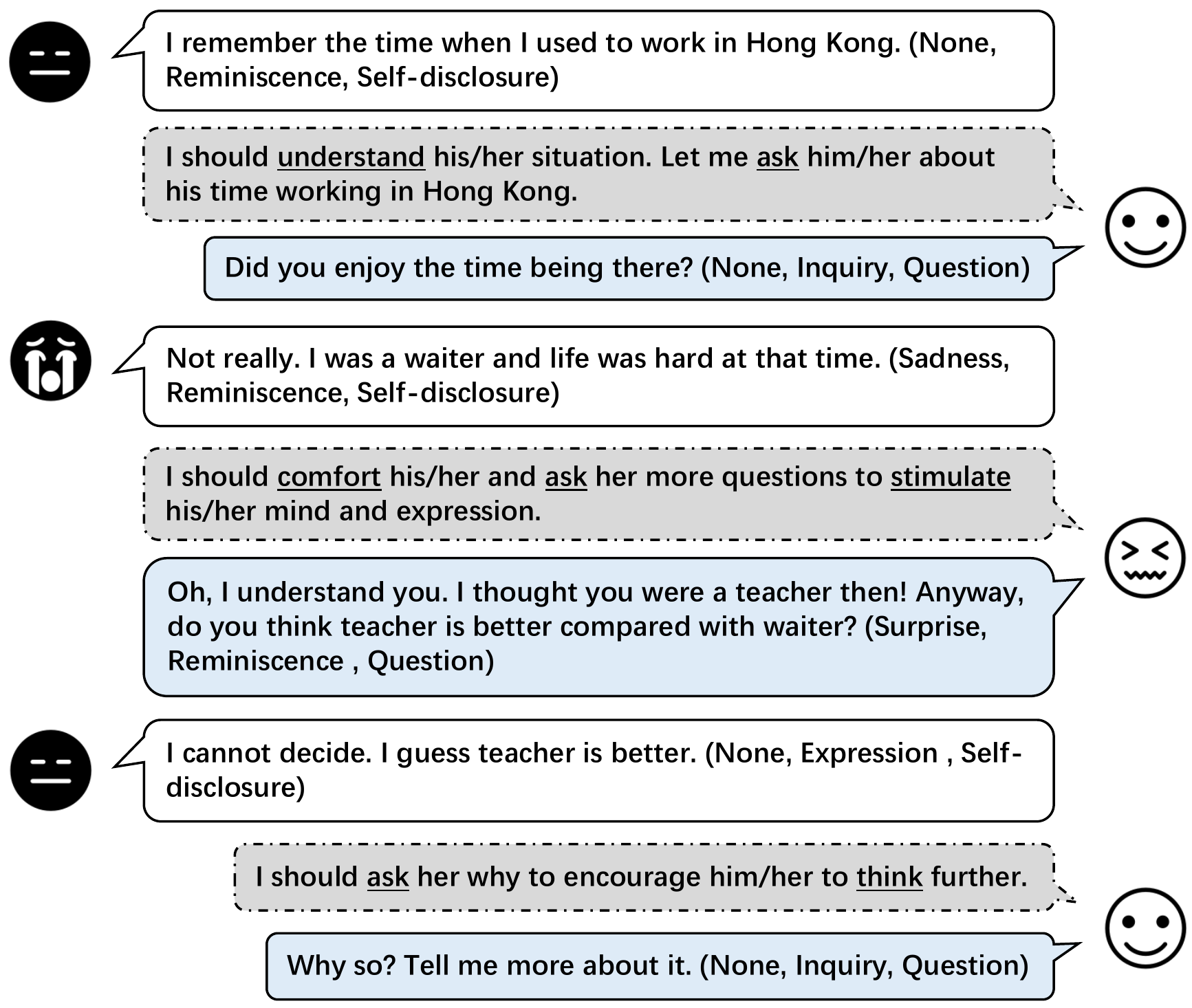}
    \caption{An example of a Chinese CS-based dialogue from the CSConv dataset (translated from Chinese to English), being provided to the elders (left) by the therapist (right). The emotion classification, CS principle, support strategy are marked in the parentheses after the utterances. The underline highlight the emotion words and keywords.} 
    \label{fig:1}
\end{figure}
Dialogue systems have enjoyed rapid progress in recent years, through communication with humans to satisfy diverse needs \citep{DBLP:journals/corr/abs-2106-01144,kann2022open}. Cognition stimulation of elders is a critical psychological therapy where dialogue systems serve as effective tools for restoring the cognition of older adults \citep{de2014beneficial, park2019effects, tokunaga2021dialogue}. 

Some studies have shown that chit-chat can help older people with cognitive restoration \cite{van2010designing, garcia2022usefulness}. 
Meanwhile, several studies have shown that emotional support is beneficial for maintaining or even increasing cognitive function in elders \cite{ellwardt2013does, liu2020emotional, sharma-etal-2020-computational}. Nonetheless, there remains an open question on how to introduce emotional support and CS principles simultaneously into chit-chat dialogue systems to provide cognitive recovery training for elders with cognitive impairment.

One main obstacle to building cognitive dialogue is the lack of training corpora, especially in the Chinese language. Therefore, we first construct a Chinese \textbf{CS Conv}ersation (\textbf{CSConv}) dataset, containing about 2.6K groups of dialogue data where each utterance is annotated with three types of labels, i.e., CS principle, emotional labels, and  emotional support strategy labels. To generate open-ended responses with emotional support strategies, we propose a multi-source knowledge fusion in a Chinese \textbf{CS D}ialogue (\textbf{CSD}) system. We use Jiagu\footnote{\url{https://github.com/ownthink/Jiagu}}, a Chinese NLP toolkit, to extract emotional words and keywords to form knowledge source and progressively mask the extracted knowledge on the encoder side, to increase the generalizability of the model. 
Meanwhile, we adopt Chinese EmoBank \citep{lee2022chinese} to calculate the weight value of each word in the utterance, so that the model pays more attention to words with high values. 
By introducing multiple sources of external knowledge, we greatly enrich the content of the conversation. 
Moreover, we judge the content and emotions that elders express which is critical to generate satisfied responses, matching them with the cognitive therapeutic principles, and coming up with corresponding supporting strategies. 
At last, we design a multi-source interactive mechanism so that emotional support strategies and cognitive stimulation therapies can be reasonably combined to generate responses benefiting to mental health.
Figure~\ref{fig:1} shows an example of a conversation with an elder based on the CS principle. 

In summary, our contributions are as follows: (1) We construct a Chinese CS-based conversation dataset to facilitate the following research; (2) We propose a progressive mask method for encoder modules, which enhances the generalizability on emotional knowledge and the applicability of the therapeutic conversations with elders; 
(3) We design a multi-source interactive method to model the interaction among encoder modules, decoder modules and external knowledge; 
(4) We conduct extensive experiments to demonstrate the effectiveness of the proposed CSD.

\section{Dataset}

\subsection{Data Collection}
There is no publicly available CS-based Chinese conversation dataset to enable a cognitively restorative dialogue system for elders with cognitive impairment. We introduce a Chinese one-to-one open-domain \textbf{CS Conv}ersation dataset, (\textbf{CSConv}), which is collected and created via cognitive stimulation therapy videos and handbook\footnote{\url{https://www.brainlive.socialwork.hku.hk/}}, and the ratio of conversation data from videos to those from the handbook is approximately 2:1. 

As high-quality conversation examples are needed for building Chinese CS-based dialogue system, our efforts include the following. (1) The videos are Cantonese. We first translate the Cantonese conversations in the videos into Mandarin Chinese, in a format suitable for CS model training. (2) We make Mandarin conversations artificially based on the eighteen CS principles in the handbook. (3) We clean the dataset based on rules (e.g., truncating excessively long utterances, removing the multiple consecutive symbols in the utterance). (4) We manually annotate whether each utterance is spoken by the SPEAKER or the LISTENER (SPEAKER for elder, LISTENER for smart speaker or health care worker). (5) We use BERT-based text classification models to annotate the emotion label, strategy label, CS label of each utterance, and then conduct manual review and modification. (6) All the data are professionally produced and reviewed twice. (7) We test our CSConv dataset on some text classification models and text generation models, which can directly reflect the performance differences between models.

\begin{table}[h]\small
\centering
\begin{tabular}{lp{5cm}lp{5cm}}
\toprule
\textbf{CS Labels} & \textbf{Explanation}\\
\midrule 
\textbf{None} & Neutral\\
\textbf{Inquiry} & Ask questions for information or open-domain questions\\
\textbf{Respect} & Be respectful or use a set pattern when talking to older people\\
\textbf{Reminiscence} & Remember things elders did when elders were a child, as well as things elders did before and personal information\\
\textbf{Expression} & Improve elders language skills and expression\\
\textbf{Enjoyment} & To have fun in conversation or to enjoy something\\
\textbf{Comfort} & Comfort the elderly to some extent\\
\bottomrule 
\end{tabular}
\caption{CS Labels and their interpretation.}
\label{TableA}
\end{table}

The CSConv dataset consists of about three thousand conversations, separated by blank rows. Each line in each conversation represents the utterance of SPEAKER or LISTENER, and SPEAKER and LISTENER’s utterances alternate. The format of each line is: SPEAKER/LISTENER utterance + <CS> + CS label + <EMO> + emotion label + <strategy> + strategy label, where <CS> is the separator of CS label and SPEAKER/LISTENER utterance; <EMO> is the separator of CS label and emotion label; <Strategy> is the separator of emotion label and strategy label. There are eight types of emotional labels, namely none, disgust, sadness, fear, surprise, like, happiness and anger. There are nine strategies (i.e., None, Question, Reflection of Feelings, Self-disclosure, Providing Suggestions, Information, Others),  which are similar to the strategies in \cite{DBLP:journals/corr/abs-2106-01144}. There are seven types of CS labels. Table~\ref{TableA} shows the name of explanation of each CS label.

\subsection{Data Statistics}
Statistics of the CSConv dataset are given in Table~\ref{TableB}. The number and proportion of CS labels, emotion labels and strategy labels are shown in Table~\ref{TableC}.

\begin{table}[h]\small
\centering
\begin{tabular}{lc}
\toprule
\textbf{Categories} & \textbf{Number}\\
\midrule
\textbf{Conversations} & 2643\\
\textbf{Utterances} & 16845\\
SPEAKER Utterances & 8363\\
LISTENER Utterances & 8480\\
Average token per conversation & 60.39\\
Average utterance per conversation & 6.37\\
Average token per utterance & 9.48\\
\bottomrule
\end{tabular}
\caption{Statistics of the CSConv dataset.}
\label{TableB}
\end{table}

\begin{table}[h]\small
\centering
\begin{tabular}{lcc}
\toprule
\textbf{CS Labels} & \textbf{Number} & \textbf{Proportion}\\
\midrule
\textbf{None} & 5296 & 31.44\\
\textbf{Inquiry} & 4156 & 24.67\\
\textbf{Respect} & 2134 & 12.70\\
\textbf{Reminiscence} & 464 & 2.76\\
\textbf{Expression} & 2651 & 15.74\\
\textbf{Enjoyment} & 1862 & 11.05\\
\textbf{Comfort} & 281 & 1.67\\
\toprule
\textbf{Emotion Labels} & \textbf{Number} & \textbf{Proportion}\\
\midrule
\textbf{None} & 12060 & 71.60\\
\textbf{Disgust} & 273 & 1.62\\
\textbf{Sadness} & 629 & 3.74\\
\textbf{Fear} & 62 & 0.37\\
\textbf{Surprise} & 355 & 2.11\\
\textbf{Like} & 1317 & 7.82\\
\textbf{Happiness} & 1954 & 11.60\\
\textbf{Anger} & 193 & 1.15\\
\toprule
\textbf{Strategy Labels} & \textbf{Number} & \textbf{Proportion}\\
\midrule
\textbf{None} & 7060 & 41.92\\
\textbf{Question} & 4195 & 24.91\\
%\textbf{Restatement or paraphrasing} & 325 & 0.02\\
\textbf{Reflection of feelings} & 293 & 17.40\\
\textbf{Self-disclosure} & 3022 & 17.94\\
%\textbf{Affirmation and reassurance} & 36 & 0.002\\
\textbf{Providing suggestions} & 262 & 1.56\\
\textbf{Information} & 819 & 4.86\\
\textbf{Others} & 1190 & 7.07\\
\bottomrule
\end{tabular}
\caption{Number and proportion of CS, emotion, strategy labels.}
\label{TableC}
\end{table}

\section{Method}

\subsection{Overview}

\begin{figure*}
\centering
	\includegraphics[scale=.5]{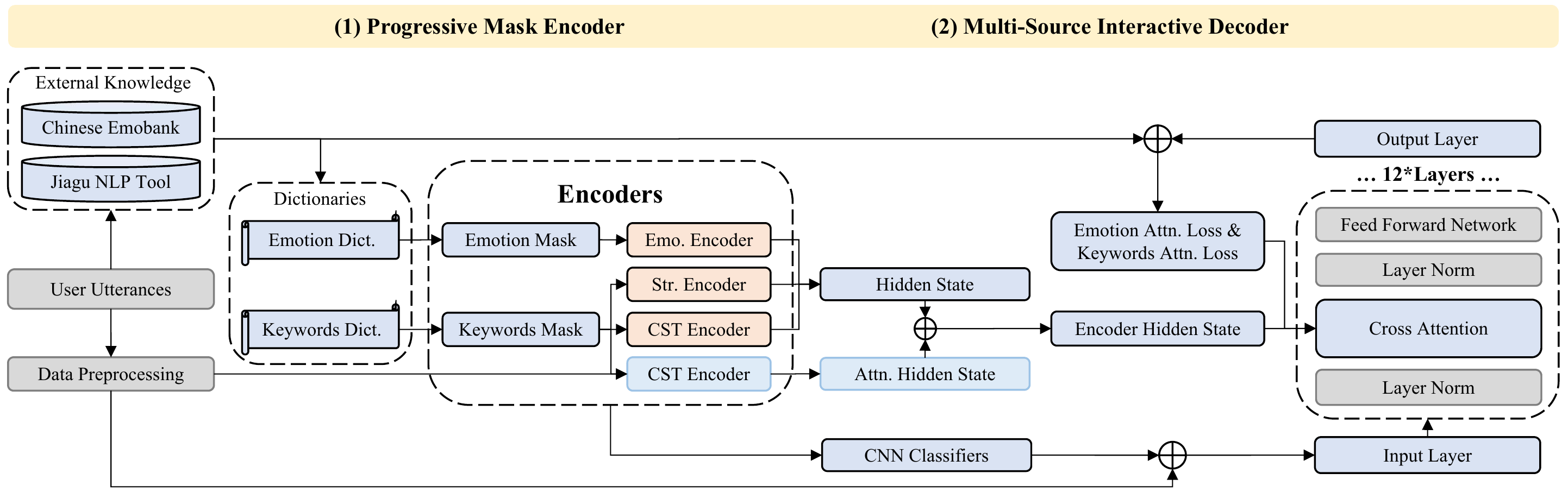}
\caption{Overall architecture of CSD.}
\label{fig:2}
\end{figure*}

Figure~\ref{fig:2} gives an overview of our Chinese CSD architecture, which consists of two stages: (1) Progressive mask encoder; (2) Multi-source interactive decoder. The first stage is divided into two modules: progressive mask encoder for context training and encoders for text classification. 

\subsection{Progressive Mask Encoder}
\textbf{Progressive Mask Encoder for Context Training.} Like the traditional BERT pre-training task, in order to better represent information of the utterances and evaluate the Next Sentence Prediction (NSP) task, the utterances of the SPEAKER and LISTENER are used to generate three types of embeddings \cite{vaswani2017attention}, namely word embedding, position embedding and segment embedding.

During training, the encoder randomly masks tokens to improve generalizability. We first use Jiagu's sentiment analysis function to extract entities (i.e., one and multiple words) and sentences with positive or negative values generated by Jiagu greater than the $\lambda_{emo}$ threshold, and Jiagu's keyword extraction function to extract keywords in the utterances. Eventually, emotion and keyword dictionaries are constructed. Through the emotion and keyword dictionaries, the data during training is masked in pre-defined proportions. As the training progresses, the span of a single mask gradually increases (i.e., from one word to multiple words, and finally to a sentence), the ratios of masking one-word entities, two-word entities, three-word entities, four-word entities and sentences are $\lambda_1$, $\lambda_2$, $\lambda_3$, $\lambda_4$ and $\lambda_5$, respectively. In order to further improve the encoder's generalization through the progressive mask method, we retain a certain proportion of the traditional BERT mask method. To be more specific, 5\% of the entities in the utterances are randomly masked, of which 80\% proceed mask processing, 10\% proceed random replacement processing, and 10\% remain unchanged. 

After the progressive mask operation, encoders are used to encode context information for the utterances (i.e., context learning) and finally the pre-trained models are obtained.

Encoders of context training based on the emotion dictionary are used for utterance emotion classification. Encoders based on the keyword dictionary are used to classify the CS principle and support strategy of the utterances. 

\textbf{Encoders for Text Classification}. A multi-turn dialogue context consists of $M$ utterances emitted by SPEAKER and LISTENER in turn. The context $\mathcal{U}$ refers to the sequence of utterance, i.e., $\mathcal{U}=[U_1, ... , U_M]$. Following \cite{DBLP:journals/corr/abs-1908-07687}, we flat $\mathcal{U}$ into a token sequence and insert a CLS token at the start of the token sentence, i.e., $\mathcal{U}=[\rm{CLS}, x_1, ... , x_m]$. 

\begin{equation}
\rm{h}_i =\rm{LN}(x_i^{l-1}+\rm{MHAtt}(x_i^{l-1}))
\end{equation}
\begin{equation}
\rm{\widetilde{x}}_i^l = \rm{LN}(h_i+\rm{FFN}(h_i))
\end{equation}

where LN is the layer normalization proposed by \cite{ba2016layer}. MHAtt is multi-head attention, which runs through an attention mechanism several times in parallel \cite{vaswani2017attention}. FFN is a two-layer feed-forward network with ReLU as the hidden activation function. The encoder contains $\rm{l}$ layers. $\rm{h}_i$ is the hidden state of the i-th token and $\rm{\widetilde{x}}_i^l$ is the embedding with context of the i-th token at the l layer. The obtained context representations are denoted as $\mathbf{C_u} = [\rm{\widetilde{x}_0}, ... , \rm{\widetilde{x}_m}]$. Let $\rm{l}_{cs}$ be the label of the CS classification result, i.e.,
\begin{equation}
\rm{l}_{cs} = \rm{CNN}(\mathbf{C_u})
\end{equation}
where CNN is a TextCNN classifier \cite{kim-2014-convolutional} 
with convolution kernel sizes (2,3,4) and 256 convolution kernels. Similarly, %to the above implementation, 
$\rm{l}_{emo}$ and $\rm{l}_{str}$ are obtained, representing the labels of the emotion classification result and the strategy classification result, respectively. 

\subsection{Multi-Source Interactive Decoder}
In the decoder generation module, we further insert a SEP token at the end of every utterance in order to distinguish the utterances between SPEAKER and LISTENER in multiple rounds of conversation, i.e., $\mathcal{U}=[\rm{CLS}, x_1, ... , x_m, \rm{SEP}]$.  

In order to generate responses more suitable for our scenario, encoders, external knowledge and decoder interact in three aspects: (1) input layer; (2) cross-attention mechanism; (3) attention loss.

\textbf{Input Layer.} We take the CS label $\rm{l_{cs}}$, emotional label $\rm{l_{emo}}$, and strategy label $\rm{l_{str}}$ that encoder classification models generate as three tokens ($t_{emo}$, $t_{cs}$, $t_{str}$)
%in the vocabulary 
and append them at the end of each utterance. We can then obtain decoder input tokens $\mathcal{Y}=[y_1, ... , y_j, t_{emo}, t_{cs}, t_{str}]$. To represent sentences and knowledge, we first use a word embedding layer, a positional embedding layer to convert each token into vectors \cite{vaswani2017attention}, i.e., $\mathbf{E}_W(y_j) \in \mathbb{R}^d$, $\mathbf{E}_P(y_j) \in \mathbb{R}^d$, where $\textit{d}$ is the dimensionality of embeddings. 
$\mathbf{y}_j$ is computed as follows:
%Our context embedding 
$[y_1, ... , y_j, t_{emo}, t_{cs}, t_{str}]$ is the composition of two types of embeddings.
%, where $\widetilde{\mathbf{y}}_j$ is computed as follows:
\iffalse
\begin{equation}
\widetilde{\mathbf{y}}_j =\mathbf{E}_W(y_j) + \mathbf{E}_P(y_j) + \mathbf{E}_D(y_j)
\end{equation}
\fi

\textbf{Cross-Attention Mechanism.} We first train an extra encoder that flattens the input data (the format of the data is the same as that of the decoder input), and get the corresponding hidden states $\rm{he_j}$: 
\begin{equation}
\rm{he}_j =\rm{LN}(y_j^{l-1}+\rm{MHAtt}(y_j^{l-1}))
\end{equation}
In order to more reasonably embed the representation of SPEAKER/LISTENR's utterances generated by encoders into the decoder through cross-attention mechanism, 
we extract the hidden states 
from the encoder classification models to replace the hidden states of the %
labels position ($\rm{he}_{emo}$, $\rm{he}_{cs}$, $\rm{he}_{str}$) generated by extra encoder, forming new encoder hidden states embedded in the cross attention of decoder.

\textbf{Attention Loss.} Since humans naturally pay extra attention to emotional support and CS information during a conversation, we enforce an emotional attention loss and keyword attention loss in order to focus on those words with higher emotion intensity values and keyword intensity values. Emotional intensity values and keyword intensity values are obtained from Chinese Emobank and Jiagu, respectively.

To highlight emotional information, we compute emotion intensity values for dialogue words and external concepts $y_j$:
\begin{equation}
\eta_{emo}(y_j) = \frac{(V_a(y_j)+A_r(y_j))-2*\rm{R_{min}}}{\rm{R_{max}}-\rm{R_{min}}}
\end{equation}
where $V_a(y_j)$ and $A_r(y_j)$ denote the mean values of valence and arousal dimensions of word $y_j$, respectively. $\rm{R_{min}}$ and $\rm{R_{max}}$ represent the minimal and maximal values of the value range, %\cwu{explain what $\rm{S_c}$ is}, %the value range
respectively. If $y_j$ is not in Chinese EmoBank, $\eta_{emo}(y_j)$ will be set to 0.

To highlight keyword information, keyword intensity values for dialogue words $y_j$ are used based on Jiagu's keyword extraction function:
\begin{equation}
\eta_{kw}(y_j) = \rm{softmax}(y_j)
\end{equation}
where the softmax operation calculates a probability for every word and the probabilities of all the words add up to one.

Emotion loss $\mathcal{L}_{emo}$ and keywords loss $\mathcal{L}_{kw}$ are calculated using Mean Square Error (MSE).

\begin{equation}
\mathcal{L}_{emo} = \frac{1}{e}\times\sum_{i=1}^e(\eta_{emo}(y_j) - a_j)^{2}
\end{equation}
\begin{equation}
\mathcal{L}_{kw} = \frac{1}{e}\times\sum_{i=1}^e(\eta_{kw}(y_j) - a_j)^{2}
\end{equation}

\noindent where $a_j$ is the attention weight of each word in the utterance calculated by the attention output tensors.

When the model generates the response, we use a sampling method to generate the next $j$-th token %\cwu{j tokens or j-th token?}.
Given $\mathcal{U}$ and tokens $t_{emo}$, $t_{cs}$ and $t_{str}$, our multi-source interactive decoder aims to generate a $n$-length response $\mathcal{Y}=\{y_1, ... , y_n\}$ through maximizing the probability $\rm{P}(\mathcal{Y}|\mathcal{U}, t_{emo}, t_{cs}, t_{str})=\prod_{n=1}^N \rm{P}(y_n|y_{<n}, \mathcal{U}, t_{emo}, t_{cs}, t_{str})$.

Like most dialogue generation tasks, standard maximum likelihood estimator (MLE) is used as the optimization objective:
\begin{equation}
\mathcal{L}_{gen} = -\rm{log}(\rm{P}(\mathcal{Y}|\mathcal{U}, t_{emo}, t_{cs}, t_{str}))
\end{equation}

Eventually, a joint loss function is defined to jointly minimize the emotion attention loss (Eq. 7), the keywords attention loss (Eq. 8) and the generation loss (Eq. 9) as follows:
\begin{equation}
    \mathcal{L} = \gamma_1 * \mathcal{L}_{gen} + \gamma_2 * \mathcal{L}_{emo} + \gamma_3 * \mathcal{L}_{kw}
\end{equation}
where $\gamma_1$, $\gamma_2$ and $\gamma_3$ are hyper-parameters.

\subsection{Training}
%During this task, 
We divide training into three phases as follows: (1) Encoders are used for context training based on the progressive mask method. Two pre-trained encoder models are trained based on sentiment dictionary and keyword dictionary, respectively. (2) CS classification and strategy classification tasks are realized on the basis of the encoder trained according to the keyword dictionary. The task of emotion classification is realized based on the encoder trained according to the emotion dictionary. (3) We use the flatten data as the training data of the encoder, %We then flat the data entered \cwu{produced?} by the encoder, 
making the batch size and input data consistent with the decoder. Then the hidden state of the last layer of the encoder is interacted with the decoder through the cross attention mechanism.

\section{Experiments}

\subsection{Implementation Details}
We conduct experiments on the CSConv dataset. For the encoder module of the CSD, the pre-trained model is bert-base-chinese\footnote{\url{https://huggingface.co/bert-base-chinese}}, and the decoder module is gpt2-chinese-cluecorpussmall \cite{zhao2019uer}. Most of the hyperparameters are the same as those in decoder chitchat\footnote{\url{https://github.com/yangjianxin1/GPT2-chitchat}}. In the progressive mask encoder trained based on the keyword dictionary, the ratios of masked entities and sentences (i.e., $\lambda_1$, $\lambda_2$, $\lambda_3$, $\lambda_4$ and $\lambda_5$) are set as 0.9, 0.9, 0.9, 0.9 and 0.4, respectively. Based on the emotion dictionary, %the ratios \cwu{what are the ratios?} 
$\lambda_1$, $\lambda_2$, $\lambda_3$, $\lambda_4$ and $\lambda_5$ are set as 0.5, 0.5, 0.4, 0.3 and 0.2, respectively. Loss weights, namely $\gamma_1$, $\gamma_2$ and $\gamma_3$, are set as 1, 0.5 and 0.5, respectively. We implement all models with PyTorch \cite{paszke2019pytorch} on four NVIDIA A100 GPUs, and train the models using AdamW optimizer \cite{loshchilov2017decoupled} with a batch size of 4. We vary the learning rate during training following \cite{vaswani2017attention}. For inference, we set the temperature as 0.7, top-k as 8 and top-p as 0.5. The training time for the encoder of the CSD is about 2 minutes and that for the decoder is about 33 minutes. In testing different models, we use NLTK packages to compute the Bleu metric and bert-score package to compute BERTScore. We set the smooth function of NLTK to method 7, and the model used in computing the bert-score is bert-base-chinese.

\begin{table*}\small
\centering
\begin{tabular}{l|c|c|c}
\toprule
\textbf{Models} & \textbf{CS Accuracy} & \textbf{Emotion Accuracy} & \textbf{Strategy Accuracy}\\
\midrule 
%TextCNN &  &  & \\
%TextRCNN &  &  & \\
Transformer & 83.67 & 85.10 & 91.63\\
BERT & 85.71 & 87.76 & 94.49\\
%ERNIE & & \\
BERT+CNN & 84.90 & 87.35 & 94.29\\
%BERT+RCNN & 86.94 & 87.96 & 93.06\\
%BERT+DPCNN & 86.94 &  & 90.02 \\
\midrule
\textbf{CSD} & \textbf{87.14} & 88.37 & \textbf{94.69}\\
\textbf{CSD (+CNN)} & 85.92 & \textbf{88.57} & 94.08 \\
%\textbf{CSBERT+RCNN} & 86.73 & \textbf{88.57} & 93.88 \\
\bottomrule
\end{tabular}
\caption{Evaluation results between baselines and the encoder module of our CSD.}
\label{TableD}
\end{table*}

\begin{table*}\small
\centering
\begin{tabular}{l|c|c|c|c|c|c|c|c}
\toprule
\textbf{Models/Products} & \textbf{Bleu-2} & \textbf{Bleu-4} & \textbf{BERTScore} & \textbf{Distinct-1} & \textbf{Distinct-2} & \textbf{Empathy} & \textbf{Support} & \textbf{Fluency} \\
\midrule
$\text{CDialGPT}_{\textrm{base}}$ & 17.55 & 6.22 & 57.70 & 8.61 & 29.34 & 3.10 & 3.11 & 3.20 \\
%CDialGPT2_{\rm{base}} & 13.73 & 4.93 & 57.33 & 7.82 & 25.91 &&&\\
$\text{CDialGPT}_{\textrm{large}}$ & 15.05 & 5.47 & 57.81 & \textbf{9.61} & \textbf{32.62} & 3.17 & 3.19 & 3.17 \\
GPT2-chitchat & 34.61 & 21.04 & 66.37 & 5.29 & 17.85 & 3.31 & 3.37 & 3.33 \\
Distil-cluecorpussmall & 39.94 & 25.30 & 69.41 & 6.44 & 22.47 & 3.27 & 3.31 & 3.29 \\
Cluecorpussmall & 41.04 & 26.59 & 68.65 & 6.79 & 23.75 & 3.39 & 3.32 & 3.39  \\
%Ownthink Robot &&&&&&&& \\
\midrule
\textbf{CSD} & \textbf{45.53} & \textbf{30.90} & \textbf{74.61} & 6.90 & 27.04 & \textbf{3.61} & \textbf{3.49} & \textbf{3.57} \\
\bottomrule
\end{tabular}
\caption{Evaluation results between baselines and our CSD. The first five metrics are automatic metrics, while the last three metrics are human metrics. \textbf{Bold face} indicates leading results in terms of the corresponding metric.}
\label{TableE}
\end{table*}

\subsection{Automatic Evaluation}
For encoder classification, to evaluate the model at the emotional level, we adopt \textbf{Emotion Accuracy} as the evaluation metric %\cwu{why it is called agreement?}
between the ground truth emotion labels and the predicted
emotion labels. \textbf{CS Accuracy} and \textbf{Strategy Accuracy} are similar evaluation metrics to emotion accuracy.

For decoder generation, we employ \textbf{BLEU} \cite{papineni-etal-2002-bleu}, an algorithm for evaluating the text quality, as the metric. Since BLEU cannot perfectly reflect the quality of generated results \cite{liu-etal-2016-evaluate}, we adopt \textbf{BERTScore} \cite{zhang2019bertscore} to compare the similarity between embeddings of a generated sentence and the reference sentence. 
\textbf{Distinct-1} / \textbf{Distinct-2} \cite{li-etal-2016-diversity} is the proportion of the distinct uni / bi-grams in all the generated results, that indicate the diversity.

\subsection{Human Evaluation}

To qualitatively examine model performance, 
we also conduct human evaluations. We sample some dialogues from the CSD and the baselines. We find 6 elders and their relatives to evaluate the responses generated by different models. All models are evaluated in terms of \textbf{Empathy}, \textbf{Support} and \textbf{Fluency}. Empathy measures whether LISTENER understands SPEAKER's feelings. Support measures whether LISTENER gives SPEAKER reasonable help and comfort. Fluency measures the grammatical correctness and readability of the SPEAKER’s responses. Each metric is rated on five-scale, where 1, 3 and 5 indicate unacceptable, moderate and excellent performance, respectively.

\subsection{Baselines for Comparison}

We conduct extensive experiments to compare the encoder module of the CSD against the following representative baselines: (1) \textbf{Transformer} \cite{vaswani2017attention}: A transformer-based encoder-decoder model. (2) \textbf{BERT} \cite{kenton2019bert}: BERT is a context-aware encoder,
%\cwu{not clear what `superimposed from encoder' means}, 
and is good at processing downstream tasks, like classification. 
(3) \textbf{BERT+CNN}\footnote{\url{https://github.com/649453932/Bert-Chinese-Text-Classification-Pytorch}}: The model is the embedding with contextual meaning output by BERT, which is input into a CNN classifier for classification.

%In decoder generation part, 
We conduct extensive experiments to compare the decoder generation 
 module of CSD against the following representative baselines: (1) \textbf{CDialGPT-base} \cite{wang2020chinese}: The model is a 12-layer GPT model trained on the LCCC-base dataset. %(5) \textbf{CDialGPT2-base} \cite{wang2020chinese}: The model is a 12-layer GPT2 model trained on LCCC-base dataset. 
(2) \textbf{CDialGPT-large} \cite{wang2020chinese}: The model is a 12-layer GPT model trained on the LCCC-large dataset. (3) \textbf{GPT2-chitchat}\footnote{\url{https://github.com/yangjianxin1/GPT2-chitchat}}: The model is a 10-layer GPT-2 trained on 500,000 chitchat corpus. (4) \textbf{Distil-cluecorpussmall} \cite{radford2019language, zhao2019uer}: The model is a 6-layer GPT-2 trained on the CLUECorpusSmall \cite{CLUECorpus2020} corpus. (5) \textbf{Cluecorpussmall} \cite{radford2019language, zhao2019uer}: The model is a 12-layer GPT-2 trained on the CLUECorpusSmall corpus.

To better analyze the influence of different components in the CSD, we also conduct an ablation study as follows: (1) \textbf{w/o $\rm{NM}$}: The CSD model uses only traditional BERT instead of BERT trained using the progressive mask method. (2) \textbf{w/o $\rm{IL}$}: The CSD model only splices three classification result labels after utterance as the train data. (3) \textbf{w/o $\rm{CA}$}: The CSD model only interacts with encoder through the cross-attention mechanism. (4) \textbf{w/o $\rm{AL}$}: The CSD model only adds the attention loss to embed external knowledge.

\begin{table}[h]\small
\centering
\begin{tabular}{l|c|c|c|c}
\toprule
\textbf{Models} & \textbf{Bleu-2} & \textbf{Bleu-4} & \textbf{BERTScore} & \textbf{Distinct-2}\\
\midrule
CSD & \textbf{45.53} & \textbf{30.90} & \textbf{74.61} & 27.04 \\
w/o $\rm{NM}$ & 44.75 & 30.42 & 74.27 & 26.77  \\
w/o $\rm{IL}$ & 42.88 & 30.53 & 73.22 & 22.71  \\
w/o $\rm{CA}$ & 43.39 & 28.73 & 72.79 & \textbf{29.54} \\
w/o $\rm{AL}$ & 43.66 & 28.91 & 70.97 & 23.20 \\
\bottomrule
\end{tabular}
\caption{Ablation test of different components.}
\label{TableG}
\end{table}

\begin{table}[h]\small
\centering
\begin{tabular}{l|c|c|c}
\toprule
\textbf{Models} & \textbf{Win} & \textbf{Loss} & \textbf{Tie}\\
\midrule
CSD vs $\text{CDialGPT}_{\rm{base}}$ & 69.0 & 20.7 & 10.3\\
%Ours vs CDialGPT2_{\rm{base}} & &&\\
CSD vs $\text{CDialGPT}_{\rm{large}}$ & 65.5 & 20.7 & 13.8\\
CSD vs GPT2-chitchat & 55.2 & 17.2 & 27.6 \\
CSD vs Distil-cluecorpussmall & 48.3 & 27.6 & 24.1 \\
CSD vs Cluecorpussmall & 41.4 & 31.0 & 27.6 \\
\bottomrule
\end{tabular}
\caption{Result of human A/B test.}
\label{TableH}
\end{table}

\subsection{Experimental Results and Analysis}
\textbf{Automatic evaluations.} In Table~\ref{TableD}, we observe that the encoder module of the CSD is better than the other baselines in CS, emotion, support strategy recognition accuracy. In Table~\ref{TableE}, we observe that the CSD outperforms strong baselines in terms of Bleu and BERTScore. %Because CSD models the therapy principle and support \cwu{what?} strategy with fewer variations
Because CSD models extensive therapy principle and emotional support strategy and there is less language diversity associated with therapy principle and emotional support strategy, the diversity of response is weaker than that of $\rm{CDialGPT_{base}}$ and $\rm{CDialGPT_{large}}$.

We also perform an ablation study for better understanding the contributions of the main modules of the CSD model. As shown in Table~\ref{TableG}, CSD outperforms all other models (w/o NM, w/o IL, w/o CA, w/o AL) in Bleu and BERTScore. However, due to CS principle and emotional support strategy intervening in the generation of decoders, %\cwu{not clear what deepening of xx means}, 
the diversity of response
%\cwu{response} 
generation decreases. Only the case of w/o CA model involving a small number of CS and support strategies achieves high diversity of generated responses.

\textbf{Human evaluation.} Table~\ref{TableE} illustrates that CSD obtains the best performance on Empathy, Support and Fluency scores. Additionally, we carry out pairwise response comparison to directly compare the dialogue quality gains in Table~\ref{TableH}. The results confirm that the responses from CSD are more preferred by human judges.

\subsection{External Knowledge Analysis}

\begin{table*}[h]\small
\centering
\begin{tabular}{l|l}
\toprule
%\textbf{\makecell[l]{ Labels }} & \makecell[l]{ \textbf{Expression, None, Self-disclosure} } \\
\textbf{\makecell[l]{ History and Labels }} &  \makecell[l]{ $X_{1}$: Where did you get your hair cut? (Inquiry, None, Question) \\  $X_{2}$: At the community center. (Expression, None, None) \\  $X_{3}$: Wow, is there anyone at the community center who cuts hair? (Inquiry, Surprise, Question) \\  $X_{4}$: Yes, it's very cheap, five dollars. (Expression, None, None) } \\
\midrule
\makecell[l]{ \textbf{Gold}  \\ $\text{CDialGPT}_{\rm{base}}$ \\ $\text{CDialGPT}_{\rm{large}}$ \\ GPT2-chitchat \\ Distil-cluecorpussmall \\ Cluecorpussmall \\ \textbf{CSD} } &  \makecell[l]{ \textbf{Wow, that's cheap. Who cut your hair?} \\ Well expensive! \\ Yes, it's very nice. \\ Yes, it's great! \\ How do you cut your hair? \\ I feel the same way. \\ \textbf{Wow, five dollars, that means they are actually volunteering, right?} } \\
\midrule
\midrule
%\textbf{\makecell[l]{ Labels }} & \makecell[l]{ \textbf{Expression, None, Self-disclosure} } \\
\textbf{\makecell[l]{ History and Labels }} &  \makecell[l]{ $X_{1}$: I've just heard "Nanping Evening Bells". (Expression, None, Self-disclosure)} \\
\midrule
\makecell[l]{ \textbf{Gold} \\ $\text{CDialGPT}_{\rm{base}}$ \\ $\text{CDialGPT}_{\rm{large}}$ \\ GPT2-chitchat \\ Distil-cluecorpussmall \\ Cluecorpussmall \\ \textbf{CSD} } & \makecell[l]{ \textbf{This song seems very old. Have you heard this song before?} \\ I just listened to it. \\ Ha ha, I just heard that too. \\ Have you ever heard the song "Nanping Evening Bells"? \\ Have you heard this song? \\ Do you seem to have heard this song before? \\ \textbf{Who sings this song? Why is it called "Nanping Evening Bells"?} } \\
\bottomrule
\end{tabular}
\caption{Generated responses (translated from Chinese to English) from CSD and baseline models. The emotion classification, CS principle, support strategy are labeled in the parentheses after the utterances. }
\label{TableF}
\end{table*}

We introduce external knowledge in three ways: training encoders by using external knowledge to progressively mask entities and sentences (w/o NM), intervening GPT-2 generation by classification labels (w/o IL), and paying more attention to emotional words and keywords by calculating the weight of words (w/o AL). To further investigate the impact of introduced knowledge, we test different components of CSD %, which are w/o NM, w/o IL, w/o AL, respectively.
as shown in Table~\ref{TableG}. However, the distinct metrics of these models are lower than models without embedded knowledge (w/o CA). Because w/o NM has more knowledge embedded than w/o IL and w/o AL and distinct metric of w/o NM is also significantly improved compared with w/o IL and w/o AL, it concluded that the generated response diversity decreases when little external knowledge is embedded, but with the increase of embedded knowledge, diversity of the generated response also increases.

\subsection{Case Study}
For decoder generation evaluation, Table~\ref{TableF} shows two examples generated by CSD and other baselines. In the first case, CSD generates an informative response with proper CS principle and emotional support, which stimulates thinking of the elder through implicit empathy and further questioning. However, %without external knowledge and encoder's interaction with decoder, all baselines 
baselines with only the decoder part fail to express responses with the CS principle and emotional support. In the second case, CSD generates a response with continuous questions, which further stimulates thinking of elder. Both cases show that CSD can generate responses with CS principle and emotional support.

\section{Related Work}

\subsection{Cognitive Training Dialogue System}

With the increasing popularity of NLP, dialogue systems have progressed from exploiting simple neural networks \cite{lee2016quote} to large-scale pre-trained models \cite{DBLP:journals/corr/abs-1910-00486, DBLP:journals/corr/abs-1911-00536, ni2022recent}. Currently, while English dialogue systems dominate, there also exist Chinese ones\footnote{\url{https://github.com/yangjianxin1/GPT2-chitchat}} \cite{wang2020large, zhou2021eva, gu2022eva2}. Most of these dialogue systems are for ordinary people, and there are few cognitive recovery dialogue systems for elders with cognitive impairment. Most of the existing dialogue systems for elders focus on specific functions, such as storytelling \cite{tokunaga2019cognitive, tokunaga2021dialogue}, robotic dialogue based on photos \cite{tokunaga2021dialogue}, etc. There are also dialogue systems for Metamemory therapy \cite{kim2021efficacy}. Few dialogue systems exist on cognitive stimulation \cite{navarro2018fuzzy}, let alone in Chinese. %and none in Chinese.

\subsection{Empathetic Dialogue, Emotional Support Conversation and Related Datasets}

With the rise of data-driven learning methods \cite{vaswani2017attention}, there are more and more studies on open-domain dialogue generation patterns \cite{DBLP:journals/corr/abs-1811-01241, kann2022open}. In order to generate an emotional response, many methods %of generating emotional responses 
automatically recognize the current user's emotional state through the conversation \cite{sabour2022cem, gao2021improving, kim2021perspective, shen2021constructing, DBLP:journals/corr/abs-2012-04080, DBLP:journals/corr/abs-1908-07687}.
\cite{li2019empdg} propose a multi-resolution adversarial framework which considers multi-granularity emotion factors and user feedback. \cite{li2022knowledge} propose a knowledge-aware empathetic dialogue generation method, which interferes with generation by embedding external knowledge into the Transformer model via diagrams. 
Some studies \cite{sharma-etal-2020-computational, WWW_meta} on empathetic dialogue technologies have also been applied to mental health. About dataset, EMPATHETICDIALOGUES \cite{rashkin-etal-2019-towards} is the benchmark of the empathetic dialogue datasets, but there exist very few relevant datasets in Chinese.

Different from empathetic dialogue, emotional support conversation can provide emotional support and problem solving in addition to empathetic responses \cite{DBLP:journals/corr/abs-2106-01144}. Because the field is new, there are a few studies on emotional support conversation \cite{MISC, peng2022fado, xu2022poke}. \cite{MISC} propose MISC, which is a mixed strategy-aware model integrating COMET for emotional support conversation.  
ESConv \cite{DBLP:journals/corr/abs-2106-01144} is the benchmark of the emotional support conversation datasets, but there is no Chinese emotional support conversation dataset.

\section{Conclusion and Outlook}
In this work, we construct a Chinese CS conversation dataset and propose a multi-source knowledge fusion method for CS dialogue. Experimental results show that CSD outperforms state-of-the-art models in terms of both automatic and human evaluations. Extensive experiments verify the effectiveness of the progressive mask method and the three interaction ways of multi-source interactive decoder in CSD. As for future work, we plan to construct larger datasets of Mandarin and Cantonese CS conversations to train models, and address the issue of CS principle, emotional support recognition in reference context in dialogue.

\label{sec:bibtex}

\section*{Limitations}
The current dialogue system is mainly based on deep neural network, like transformer structure, which often requires a large number of data sets for training model. However, there are still some deficiencies in our dataset. We will further label and create more dataset to train model. In addition, in order to improve the quality of dialogue, our model parameters are relatively large, which affect the speed of dialogue generation to some extent. We will explore some methods, such as knowledge distillation, to reduce model parameters to improve the speed of dialogue generation on the premise of keeping the quality of dialogue generation unchanged.

\section*{Ethics Statement}
We have sought to ethically conduct this study, including transparently communicating with data annotators 
about data use and study intent, and finding suitable elders to conduct human tests of the dialogue systems, compensating workers and elders at a reasonable hourly wage. We have obtained study approval from the ethics review board.

\section*{Acknowledgements}

We want to thank our anonymous AC and reviewers for their feedback. This work was supported in part by grants from Hong Kong Research Grants Council (RGC) under the contracts HKU 17203522 and 17207621.

\bibliography{custom}
\bibliographystyle{acl_natbib}

\end{document}